\documentclass[conference]{IEEEtran}
\makeatletter\if@twocolumn\PassOptionsToPackage{switch}{lineno}\else\fi\makeatother
\usepackage{bigdelim}

\usepackage{graphicx}
\usepackage[T1]{fontenc}
\usepackage[numbers,sort&compress]{natbib}
\usepackage{mathtools}
\usepackage{booktabs}
\usepackage{lineno,hyperref}
\usepackage{amsmath}
\usepackage{amsfonts}
\usepackage{amssymb}
\usepackage{amsmath}
\usepackage{multirow,morefloats,floatflt,cancel,tfrupee}
\usepackage{makecell}
\usepackage{tikz}
\usepackage{setspace}
\usepackage{color}
\usepackage{multirow}
\usepackage{adjustbox}
\usepackage{bbding}
\usepackage{physics}
\usepackage{colortbl}
\usepackage{xcolor}
\usepackage{pifont}
\usepackage{subcaption}
\usepackage[nointegrals]{wasysym}
\renewcommand\IEEEkeywordsname{Keywords}

\makeatletter
\def\ps@IEEEtitlepagestyle{%
  \def\@oddfoot{\mycopyrightnotice}%
  \def\@evenfoot{}%
}
\def\mycopyrightnotice{%
  {\footnotesize 979-8-3315-1127-2/24/\$31.00  \copyright 2024 IEEE \hfill}
  \gdef\mycopyrightnotice{}
}

\let\old@ps@IEEEtitlepagestyle\ps@IEEEtitlepagestyle
\def\confheader#1{%
    \def\ps@IEEEtitlepagestyle{%
        \old@ps@IEEEtitlepagestyle%
        \def\@oddhead{\strut\hfill#1\hfill\strut}%
        \def\@evenhead{\strut\hfill#1\hfill\strut}%
    }%
    \ps@headings%
}
\makeatother

\confheader{%
    \parbox{17cm}{\centering 14th International Conference on Computer and Knowledge Engineering (ICCKE 2024), November 19-20, 2024, Ferdowsi University of Mashhad}
}

\begin{document}

        \title{Enhancing Vehicle Make and Model Recognition with 3D Attention Modules}
%


\author{\IEEEauthorblockN{Narges Semiromizadeh,
Omid Nejati Manzari,
Shahriar B. Shokouhi,
Sattar Mirzakuchaki}\\
School of Electrical Engineering, Iran University of Science and Technology, Tehran, Iran}

\maketitle

\begin{abstract}

Vehicle make and model recognition (VMMR) is a crucial component of the Intelligent Transport System, garnering significant attention in recent years. VMMR has been widely utilized for detecting suspicious vehicles, monitoring urban traffic, and autonomous driving systems. The complexity of VMMR arises from the subtle visual distinctions among vehicle models and the wide variety of classes produced by manufacturers. Convolutional Neural Networks (CNNs), a prominent type of deep learning model, have been extensively employed in various computer vision tasks, including VMMR, yielding remarkable results. As VMMR is a fine-grained classification problem, it primarily faces inter-class similarity and intra-class variation challenges.
In this study, we implement an attention module to address these challenges and enhance the model's focus on critical areas containing distinguishing features. This module, which does not increase the parameters of the original model, generates three-dimensional (3-D) attention weights to refine the feature map. Our proposed model integrates the attention module into two different locations within the middle section of a convolutional model, where the feature maps from these sections offer sufficient information about the input frames without being overly detailed or overly coarse.
The performance of our proposed model, along with state-of-the-art (SOTA) convolutional and transformer-based models, was evaluated using the Stanford Cars dataset. Our proposed model achieved the highest accuracy, 90.69\%, among the compared models.
\end{abstract}


\section{Introduction}

In the field of Intelligent Transportation Systems (ITS), a novel topic pertaining to vehicle analysis is VMMR. In the automotive industry, the term "make" denotes the producer of a specific vehicle (such as Hyundai, Ford, and Toyota), whereas "model" specifies a certain kind of vehicle created by these producers (such as Azera, Corolla, and Focus). Traffic control, surveillance, traffic statistics, law enforcement, self-driving vehicles, and the detection of suspicious vehicles are just a few of the fields in which VMMR systems are extensively utilized \cite{ali2022vehicle}.

VMMR can be regarded as a kind of fine-grained classification, which aims to identify and distinguish between categories with very similar features in a classification problem. The enormous existing categories, low inter-class and high intra-class variance make VMMR a highly challenging problem. The appearance of subtle visual disparities in specific cases can make recognition extremely difficult, even for humans \cite{gayen2023two}.

Nowadays, deep learning networks are frequently used in varying computer vision tasks, one of which is image recognition. CNNs, as one of the deep learning models, have remarkably enhanced the performing of vision-related tasks due to their rich representational power \cite{he2016deep, hu2018squeeze, huang2017densely, manzari2024denunet}. Many of the proposed methods for VMMR have utilized CNN-based models and have achieved remarkable results. Transformer is also a kind of deep neural network that was first used in the field of natural language processing (NLP). Motivated by the significant achievement of the transformer in the field of NLP, transformer based network is used in many computer vision tasks including image recognition \cite{vaswani2017attention}. Since the transformer is based on the self-attention mechanism, it is suitable for extracting long-range dependencies and is more capable of extracting global features \cite{manzari2025medical}. In contrast, CNNs are more capable of extracting local features.

In human perception, attention is crucial. One key aspect of the human eye anatomy is that individuals do not seek to process an overall scene simultaneously. Rather, people utilize a subset of partial glimpses and selectively concentrate on prominent elements to more effectively grasp visual structure. This concept inspired the creation of the attention mechanism in deep learning \cite{woo2018cbam}. Since in computer vision tasks, including image recognition, both local and global features are significantly important, the attention mechanism can be employed to enhance CNNs' ability to extract global information.

In this paper, we present an enhanced network for VMMR that incorporates a simple and parameter-free attention module (SimAM). This module, which we demonstrate to be more efficient than other existing SOTA attention modules, is combined with a suitable convolutional model in an effective manner. The result is a significant increase in the model's representational power. The attention mechanism guides the model to focus on important section of an image, extracting the most information while ignoring less important areas. This reduction in the negative impact of intra-class variation and inter-class similarity instills confidence in the effectiveness of our approach.
The rest of this study is organized as bellows:

We investigate the related works in Section II. Section III explains the structure of our proposed model. In Section IV, we demonstrate the results of evaluating the performance of our proposed network and other SOTA architectures and analyze our results. Finally, we bring to a conclusion our study in Section V.

\begin{figure*}[!t]
 \centering
  \includegraphics[width=.8\textwidth]{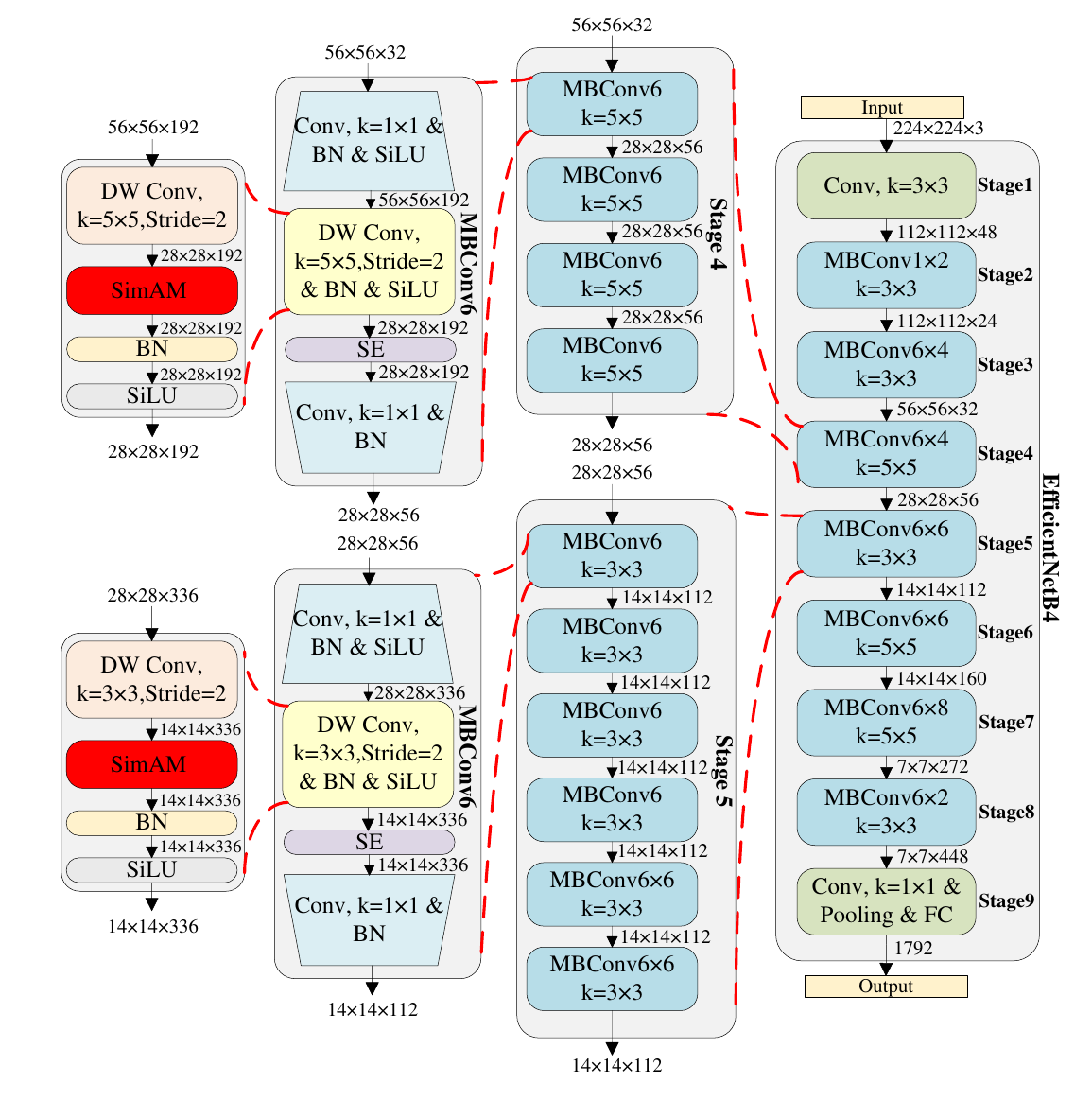}
  \caption{Network architecture of the proposed model.}\label{fig.ViT}
  \vspace{-4mm}
\end{figure*}

\section{Related Works}
In this section, several notable works on image recognition models and attention modules are briefly reviewed.

\subsection{Deep Learning Models}
\textbf{CNNs} are commonly found in a variety of computer vision tasks, including VMMR. Several studies concentrate on designing model architectures to enhance representational power of CNNs. For instance, in ResNet \cite{he2016deep}, shortcut connections are utilized, allowing the CNN to expand to hundreds of layers. Their findings reveal that raising the architecture depth can considerably strengthen the representation power of a CNN. DenseNet \cite{huang2017densely} employs dense connections across layers, allowing each layer to access the feature maps of all previous layers as inputs. This design provides several key advantages: it helps alleviate the vanishing-gradient issue, facilitates feature reuse, improves feature propagation, and notably reduces the total parameter count. Based on the depthwise separable convolutions \cite{sandler2018mobilenetv2}, proposes MobileNetV2, which is specifically designed for limited resources and mobile environments, which uses a new inverted residual structure as the building block. Using neural architecture search, \cite{tan2019efficientnet} designs a new baseline architecture and scales it up using a new compound scaling method to derive a family of models known as EfficientNets that outperform earlier CNNs in terms of efficiency and accuracy.

\textbf{Transformer} performance in the field of NLP drew the attention of the computer vision society to it, and with the introduction of ViT, Transformer was seriously incorporated into computer vision tasks. Following the ViT, other transformer-based models were proposed to improve performance in computer vision tasks. For example, Swin Transformer \cite{liu2021swin}, which functions effectively as a versatile backbone for computer vision, generates a hierarchical structure design and has linear computational cost in relation to the size of the input image. Shifted window-based self-attention, a fundamental component of Swin Transformer, is demonstrated to be successful and efficient in addressing vision challenges. CrossViT \cite{chen2021crossvit} is a dual-branch transformer architecture for extracting multi-scale features, in which a fusion approach built on cross-attention is used to effectively mix picture tokens of different patch scales and exchange information between two branches.

\textbf{hybrid} architectures combine transformers with convolutional layers to introduce locality into the transformer model. MaxViT \cite{tu2022maxvit} utilizes a hierarchical structure with multi-axis attention mechanisms for efficient local and global feature extraction. This design integrates both convolutional and self-attention layers in a novel and effective manner. The hierarchical transformer design, Pyramid ViT (PVT) \cite{wang2021pyramid}, proposes a sequential shrinking pyramid and dimensional reduction attention. PVT-V2 enhances the original PVT by incorporating three key features: firstly, an attention layer with linear complexity, secondly, overlapping patch embeddings, and finally a convolutional feed-forward network.

\subsection{Attention Modules}

Various works have introduced attention modules that can be applied to many CNNs \cite{manzari2024befunet, saadati2023dilated}. These attention modules modify feature maps, empowering the CNN to target on important areas of the image or specific and informative features and make decisions based on them. A representative work is Squeeze-and-Excitation (SE) \cite{hu2018squeeze}, which proposes channel-wise attention and models the interdependencies between the channels. This is achieved via a three-step process: squeeze (global average pooling), excitation (learning channel dependencies with fully connected layers), and scale (reweighting feature maps). Later works, CBAM \cite{woo2018cbam} combines spatial and channel-wise attention mechanisms. SRM \cite{lee2019srm} improves a CNN's representational strength by embedding styles into the feature maps. GC \cite{cao2020global} effectively captures long-range dependencies to enhance semantic comprehension. ECA \cite{wang2020eca} adopts a local cross-channel interaction method without dimensionality reduction, implemented efficiently through 1D convolution. CA \cite{hou2021coordinate} presents a new lightweight attention method, termed Coordinate Attention, designed for mobile networks. This mechanism combines the advantages of channel attention in modeling inter-channel relationships with the ability to capture long-range dependencies and maintain accurate positional information. To compute the 3-D weights, SimAM \cite{yang2021simam} designs an energy function using some established neuroscience theories.Using simple yet effective operations, such as average pooling to capture long-range dependencies and the dot product to represent cross-dimensional interactions, SIAM \cite{han2024siam} generates 3-D attention maps with minimal computational overhead.

\section{METHODOLOGY}

In this section, we present our proposed model. To provide a clearer understanding of its architecture, we first review the characteristics of some well-known attention modules. We then outline the formulation of SimAM \cite{yang2021simam} before detailing our model's structure.

\subsection{Revisiting attention modules}

Attention modules are generally embedded within each block to enhance the outputs of preceding layers. This refinement usually occurs along either the channel or spatial dimension, producing 1-D or 2-D weights and treating neurons similarly within each channel or spatial position, which may limit the capacity to learn more distinct features. Consequently, generating 3-D attention weights and assigning a unique weight to each neuron for refining the feature map within a layer is considered more effective.

Direct estimation of 3-D weights is challenging. In \cite{wang2017residual}, the use of an encoder-decoder framework for learning 3-D weights is proposed. However, this method introduces various sub-networks from lower layers to higher layers of a ResNet \cite{he2016deep}, making it difficult to integrate into other modularized networks. CBAM \cite{woo2018cbam} is another example, applying channel and spatial attention modules in sequence without directly generating 3-D weights. This two-step approach in CBAM requires substantial computation time, suggesting that 3-D weight computation should be streamlined to keep the module efficient and lightweight.

The method of weight production is another crucial aspect of attention modules. Designing the structure requires significant engineering, and attention weights are calculated using a few groundless heuristics in the majority of studies that currently exist. For instance, compared to SE \cite{hu2018squeeze}, CBAM \cite{woo2018cbam} uses global max pooling (GMP) in addition to global average pooling (GAP) to create finer attention. As shown in Table \ref{tab:attention}, attention modules are based on many common operators such as FC, Conv2D, BN, etc., and some special operators such as channel-based convolution (C1D).

SimAM can produce 3-D attention weights for a layer’s feature map, assigning a distinct weight to each neuron. This module is inspired by the concept of spatial suppression in mammalian brains, utilizing an energy function to calculate these weights. Furthermore, SimAM \cite{yang2021simam} speeds up weight computation by employing a fast closed-form solution for the energy function.

\begin{table}[ht]
    \centering
    \caption{Comparison of various attention modules based on their structural design and parameters. The operators include: channel-based convolution (C1D) or standard convolution (C2D), spatial (GAP) or channel average pooling (CAP), channel max pooling (CMP) or spatial max pooling (GMP), batch normalization (BN), layer normalization (LN), standard deviation computed along the spatial dimension (GSP), standard fully-connected (FC) or channel-wise fully-connected (CFC) layers, as well as activation functions like Softmax and ReLU. Here, kk and rr denote the filter count and reduction ratio, respectively, while CC represents the current number of feature channels.}\label{tab:attention}
    \begin{adjustbox}{width=.95\linewidth,center}
    \begin{tabular}{|c|c|c|c|}
        \hline
        Attention Modules & Operators & Parameters & Design \\
        \hline
        CBAM \cite{woo2018cbam} &  C2D, GAP, GMP, FC, ReLU, CAP, CMP, BN & $2C^2 / r + 2K^2$ & handcrafted \\
        \hline
        SE \cite{hu2018squeeze} & GAP, FC, ReLU & $2C^2 / r$ & handcrafted \\
        \hline
        GC \cite{cao2020global} & C1D, Softmax, LN, FC, ReLU & $2C^2 / r + C$ & handcrafted \\
        \hline
        SRM \cite{lee2019srm} & GAP, GSP, CFC, BN & $6C$ & handcrafted \\
        \hline
        ECA \cite{wang2020eca} & GAP, C1D & $K$ & handcrafted \\
        \hline
        SimAM \cite{yang2021simam}& GAP, $\mathcal{I}$, $\mathcal{O}$, + & $0$ & Eqn (5) \\
        \hline
    \end{tabular}
    \end{adjustbox}
\end{table}

\subsection{SimAM attention module}

In SimAM \cite{yang2021simam}, each neuron's importance is estimated to produce 3-D attention weights. In visual neuroscience, neurons that provide the most critical information often display unique patterns compared to their neighbors. Additionally, the spatial suppression phenomenon indicates that an active neuron can suppress the activity of its neighboring neurons. Neurons exhibiting strong spatial suppression effects tend to be more significant in visual processing. To identify such neurons, we can measure the linear separability between a target neuron and its surrounding neurons. This understanding leads to defining each neuron's energy function as follows:

\begin{equation}
e_n\left(w_n, b_n, p, q_i\right)=\left(p_n-\hat{n}\right)^2+\frac{1}{M-1} \sum_{i=1}^{M-1}\left(p_o-\hat{q}_i\right)^2
\end{equation}

Here, the linear transformations of \(n\) and \(q_i\) are \(\hat{n}=w_n n+b_n\) and \(\hat{q}_i=w_n q_i+b_n\), where \(n\) and \(q_i\) represent the target neuron and other neurons within a single channel of the input feature \(Z \in \mathbb{R}^{D \times H \times W}\). \(i\) indexes across the spatial dimension, and \(M=H \times W\) represents the number of neurons in that channel. \(w_n\) and \(b_n\) are the weight and bias of the transformation. When \(\hat{n}\) matches \(p_n\) and all other \(\hat{q}_i\) are \(p_o\) (where \(p_n\) and \(p_o\) are distinct values), Eqn (1) achieves its minimum value. Minimizing Eqn (1) equates to identifying the linear separability between the target neuron \(n\) and all other neurons within the same channel. By assigning binary labels (e.g., 1 and -1) to \(p_n\) and \(p_o\) and introducing a regularizer, the final energy function becomes:

\begin{equation}
\begin{aligned}
e_n\left(w_n, b_n, p, q_i\right) & =\frac{1}{M-1} \sum_{i=1}^{M-1}\left(-1-\left(w_n q_i+b_n\right)\right)^2 \\
& +\left(1-\left(w_n n+b_n\right)\right)^2+\lambda w_n^2
\end{aligned}
\end{equation}

The closed-form solution for Eqn (2) is computed as follows:

\begin{equation}
w_n=-\frac{2\left(n-\alpha_n\right)}{\left(n-\alpha_n\right)^2+2 \beta_n^2+2 \lambda}
\end{equation}

\begin{equation}
b_n=-\frac{1}{2}\left(n-\alpha_n\right) w_n
\end{equation}

\(\beta_n^2=\frac{1}{M-1} \sum_{i=1}^{M-1}\left(q_i-\alpha_n\right)^2\) and \(\alpha_n=\frac{1}{M-1} \sum_{i=1}^{M-1} q_i\) are the variance and mean calculated over all neurons except \(n\) in that channel. Assuming that all pixels of a single channel follow the same distribution, the mean and variance are calculated once and reused across all neurons in that channel, significantly reducing computational costs. Thus, the minimum energy is computed as follows:

\begin{equation}
e_n^*=\frac{4\left(\hat{\beta}^2+\lambda\right)}{(n-\hat{\alpha})^2+2 \hat{\beta}^2+2 \lambda}
\end{equation}

where \(\hat{\alpha}=\frac{1}{M} \sum_{i=1}^M q_i\) and \(\hat{\beta}^2=\frac{1}{M} \sum_{i=1}^M\left(q_i-\widehat{\alpha}\right)^2\). Eqn (5) suggests that a lower energy \(e_n^*\) indicates that the neuron \(n\) is more distinct from its neighbors, making it more important for visual processing. Consequently, each neuron's importance can be represented by \(1 / e_n^*\). The final refinement phase of the module is expressed as:

\begin{equation}
\tilde{Z}=\operatorname{sigmoid}\left(\frac{1}{F}\right) \odot Z
\end{equation}

where \(F\) aggregates all \(e_n^*\) across channel and spatial dimensions. The sigmoid function is included to constrain overly large values in \(F\) \cite{yang2021simam}.

\begin{table}[h]
\caption{EfficientNet-B4 network -- Each row describes a stage $i$ with $\mathcal{L}_{i}$ layers, with input resolution ($\hat{H}_{i} \times \hat{W}_{i}$) and output channels $\mathcal{C}_{i}$.}\label{tab:efficientnet-b4}
\begin{adjustbox}{width=.9\linewidth,center}
\begin{tabular}{|c|c|c|c|c|}
\hline
\textbf{Stage(i)} & \textbf{Operator($\mathcal{F}_{i}$)} & \textbf{Resolution($\hat{H}_{i} \times \hat{W}_{i}$)} & \textbf{\#Channels($\mathcal{C}_{i}$)} & \textbf{\#Layers($\mathcal{L}_{i}$)} \\ \hline
1 & Conv3$\times$3 & 224$\times$224 & 48 & 1 \\ \hline
2 & MBConv1, k3$\times$3 & 112$\times$112 & 24 & 2 \\ \hline
3 & MBConv6, k3$\times$3 & 112$\times$112 & 32 & 4 \\ \hline
4 & MBConv6, k5$\times$5 & 56$\times$56 & 32 & 4 \\ \hline
5 & MBConv6, k3$\times$3 & 28$\times$28 & 112 & 6 \\ \hline
6 & MBConv6, k5$\times$5 & 14$\times$14 & 160 & 6 \\ \hline
7 & MBConv6, k5$\times$5 & 14$\times$14 & 272 & 8 \\ \hline
8 & MBConv6, k3$\times$3 & 7$\times$7 & 1792 & 2 \\ \hline
9 & Conv3$\times$3 \& Pooling \& FC & 7$\times$7 & 1792 & 1 \\ \hline
\end{tabular}
\end{adjustbox}
\end{table}

\subsection{Using SimAM within EfficientNet}

SimAM is compatible with most standard CNN architectures. We select a strong convolutional model, EfficientNet-B4 \cite{tan2019efficientnet}, as the baseline model. By deploying SimAM to this baseline model, we obtain a new model called SimAM-EfficientNetB4, as shown in Figure 1. EfficientNet’s core building block is the mobile inverted bottleneck, MBConv, which is the same inverted residual block used in the structure of MobileNetV2 \cite{sandler2018mobilenetv2}. However, it not only uses 3×3 convolutions but also 5×5 convolutions and includes the squeeze-and-excitation optimization \cite{hu2018squeeze}. According to Table \ref{tab:efficientnet-b4}, which details the architecture of EfficientNet-B4, it consists of nine stages. The fourth stage has four MBConv blocks and the fifth stage has six MBConv blocks. The attention module is incorporated into a convolutional neural network at a layer where the feature map extracted holds an optimal level of information from the input frames, balancing between detail and abstraction. To this purpose, the SimAM is added to EfficientNet-B4 in two parts: 1) In the first MBConv of the fourth stage, after the 5×5 depthwise convolution (DW Conv) and before BN. The attention module receives a feature map of size 28×28×192 and outputs a modified feature map of the same size, 28×28×192. 2) In the first MBConv of the fifth stage, after the 3×3 depthwise convolution and before BN. The attention module receives a feature map of size 14×14×336 and outputs a modified feature map of the same size, 14×14×336.

 \begin{figure}[!t]
 \centering
  \includegraphics[width=\linewidth]{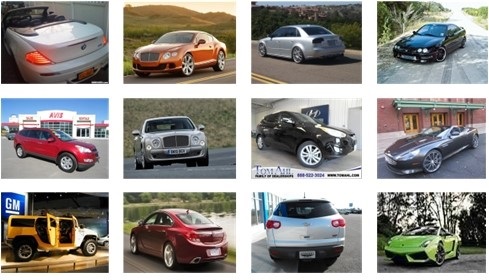}
  \caption{Some examples of vehicles in stanford car dataset.}\label{fig.dataset}
  \vspace{-4mm}
\end{figure}

\section{Experiments}

In this section, we begin by presenting the dataset utilized to train our proposed model, followed by a description of the experimental settings. Next, we perform an ablation study to demonstrate the impact of the coefficient value within the attention module of our model. Lastly, we compare our model with state-of-the-art (SOTA) models, evaluating accuracy, model parameters, and floating-point operations (FLOPs) on the target dataset.

\subsection{Dataset and Implementation details}

We conduct experiments and evaluate the effectiveness of our proposed model on a vehicle dataset. The Stanford Cars \cite{krause20133d} dataset consists of 16,185 images of cars across 196 classes, with roughly 50 by 50 divisions between training and testing data. The training set includes 8,144 images, while the testing set comprises 8,041 images, each representing one of the 196 car classes. The images are taken from different angles and provide different views of the vehicles. Samples from this dataset are shown in Figure 2.

To conduct our experiments, we used Google Colab as our coding environment, which offered a Tesla T4 GPU and 15 gigabytes of RAM. We used the PyTorch framework to implement our experiments. We trained and tested our proposed and several SOTA models on the Stanford Cars dataset, including convolutional and transformer-based models. We used the Adam optimizer to train our proposed model and all convolutional models. Additionally, a CosineAnnealingLR scheduler with a T-max of 21000 and eta-min of 1e-7 was employed to decrease the learning rate during the training process. All transformer-based models were trained using the Stochastic Gradient Descent (SGD) optimizer with a momentum of 0.9 and a weight decay of 1e-4. A StepLR scheduler was utilized to decrease the learning rate during the training process, such that the learning rate was divided by 10 after every 20 epochs. All models were trained for 80 epochs. The initial learning rate was set to 1e-3. We set the batch size for both testing and training to 32 and used the Cross-Entropy loss function. The images were resized to 224 × 224. We employed several data augmentation techniques, such as random horizontal flip, random rotation up to 15, random grayscale, and random posterization during training.

\begin{table}[h!]
    \caption{Performance evaluation of proposed model and SOTA models on stanford cars dataset.}\label{tab:stanford}
    \begin{adjustbox}{width=.9\linewidth,center}
    \begin{tabular}{l|ccc}
    \toprule
    Model & Accuracy  &  Parameters(M) & FLOPs(G) \\
    \midrule
    ResNet18 & $87.04$ & $21.39$ & $3.68$ \\
    DenseNet201 & $73.71$ & $18.47$ & $4.34$ \\
    MobileNetV2 & $88.57$ & $4.67$ & $0.59$ \\
    EfficientNet-B4 & $89.39$ & $17.9$ & $1.55$ \\
    Swin-tiny & $87.10$ & $27.67$ & $4.38$ \\
    DeiT-small& $84.44$ & $21.74$ & $3.22$ \\
    CrossViT-small & $77.75$ & $26.39$ & $4.07$ \\
    ConViT-small & $86.84$ & $27.43$ & $5.37$ \\
    PVT-V2 & $87.51$ & $22.14$ & $3.81$ \\
    MaxViT-tiny & $86.32$ & $28.64$ & $4.93$ \\ \midrule
\rowcolor{gray!10} Proposed Model & $90.69$ & $17.9$ & $1.55$ \\
    \bottomrule
    \end{tabular}
\end{adjustbox}
\end{table}

\begin{table}[h!]
\caption{Investigation of different $\lambda$ values in performance of proposed model.}\label{tab:hyper}
\begin{adjustbox}{width=.9\linewidth,center}
\begin{tabular}{|c|c|c|c|}
\toprule
\rowcolor{gray!20}  $\lambda$ & Accuracy(\%) & Parameters(M) & FLOPs(G) \\
\midrule
  7e-1 & 89.90 & $17.9$ & $1.55$ \\
\rowcolor{gray!10} 7e-2 & 89.85 & $17.9$ & $1.55$ \\
 7e-3 & 90.04 & $17.9$ & $1.55$ \\
\rowcolor{gray!10} \textbf{7e-4} & \textbf{90.69} & \textbf{17.9} & \textbf{1.55} \\
  7e-5 & 90.01 & $17.9$ & $1.55$ \\
\bottomrule
\end{tabular}
\end{adjustbox}
\end{table}

\subsection{RESULTS}

Table \ref{tab:stanford} presents a performance evaluation of various SOTA models alongside the proposed model using the Stanford Cars dataset. The comparison is based on three metrics: Accuracy, Parameters in millions, and FLOPs in billions. The proposed model achieves the highest accuracy at 90.69\%, outperforming all other models while maintaining a relatively low parameter count and FLOPs. This indicates that our model is not only more accurate but also more efficient in terms of computational resources compared to many SOTA methods.

Table \ref{tab:hyper} examines the performance metrics of the proposed model at various values of the hyperparameter $\lambda$. The proposed model exhibits its best performance with an accuracy of 90.69\% at $\lambda = 7 \times 10^{-4}$. The parameters and FLOPs remain unchanged across different $\lambda$ values, indicating that the computational efficiency and model complexity are unaffected by the variations in $\lambda$. Therefore, adjusting $\lambda$ primarily impacts the accuracy, with $\lambda = 7 \times 10^{-4}$ being the optimal value for the highest accuracy.

\section{Conclusion}
In this study, we addressed the challenges of VMMR by integrating an attention module into a convolutional model. The attention module, designed to focus on critical areas containing distinguishing features, was incorporated parameter-free, enhancing the model's ability to handle inter-class similarity and intra-class variation without increasing computational complexity.
Using the Stanford Cars dataset, our proposed model was evaluated alongside CNN and transformer-based methods. The results demonstrated that our model outperformed all other compared models. The analysis also showed that our model maintained a relatively low parameter count and FLOPs, indicating its efficiency in computational resources.
Future work could explore the application of this attention mechanism to other fine-grained classification problems and investigate its integration with more advanced neural network architectures.

\bibliographystyle{IEEEtran}

{\small
\bibliography{article}}

\end{document}